\documentclass[letterpaper]{article} 
\usepackage{aaai23}  
\usepackage{times}  
\usepackage{helvet}  
\usepackage{courier}  
\usepackage[hyphens]{url}  
\usepackage{graphicx} 
\usepackage{natbib}  
\usepackage{caption} 
\usepackage{algorithm}
\usepackage{algorithmic}

\usepackage{amsmath}
\usepackage{amssymb}
\usepackage{bbm}
\usepackage{amsmath,bm}
\usepackage{xcolor}
\usepackage{multirow}

\usepackage{booktabs}
\usepackage{pifont}
\usepackage{wrapfig}
\usepackage{newfloat}
\usepackage{listings}
\DeclareCaptionStyle{ruled}{labelfont=normalfont,labelsep=colon,strut=off} 
\floatstyle{ruled}
\newfloat{listing}{tb}{lst}{}
\floatname{listing}{Listing}
\title{Controllable Image Captioning via Prompting}
\author{
	Ning Wang,~
	Jiahao Xie,~
	Jihao Wu,~
	Mingbo Jia,~
	Linlin Li \\
}
\affiliations{
    Huawei Inc.\\
    wn6149@mail.ustc.edu.cn,~
    jh\_xie@tongji.edu.cn,~
    \{wujihao,~jiamingbo,~lynn.lilinlin\}@huawei.com 
}

\usepackage{bibentry}

\begin{document}

\maketitle

\begin{abstract}
	Despite the remarkable progress of image captioning, existing captioners typically lack the controllable capability to generate desired image captions, e.g., describing the image in a rough or detailed manner, in a factual or emotional view, etc.
	In this paper, we show that a unified model is qualified to perform well in diverse domains and freely switch among multiple styles.
	Such a controllable capability is achieved by embedding the prompt learning into the image captioning framework.
	To be specific, we design a set of prompts to fine-tune the pre-trained image captioner.
	These prompts allow the model to absorb stylized data from different domains for joint training, without performance degradation in each domain.
	Furthermore, we optimize the prompts with learnable vectors in the continuous word embedding space, avoiding the heuristic prompt engineering and meanwhile exhibiting superior performance. 
	In the inference stage, our model is able to generate desired stylized captions by choosing the corresponding prompts.
	Extensive experiments verify the controllable capability of the proposed method.
	Notably, we achieve outstanding performance on two diverse image captioning benchmarks including COCO Karpathy split and TextCaps using a unified model.
\end{abstract}

\section{Introduction}\label{section:introduction}

Image captioning is one of the fundamental tasks in computer vision, which aims to automatically generate natural and readable sentences to describe the image contents.
The last decade has witnessed the rapid progress of image captioning, thanks to the development of sophisticated visual representation learning \cite{VINVL,ViTCap}, cross-modal fusion \cite{XLAN,AoANet,OSCAR}, vision-language pre-training \cite{LEMON,BLIP,SimVLM}, etc.
Image captioning is a challenging task that requires the captioners to recognize the objects and attributes, understand their relationships, and properly organize them in the sentence.

\begin{figure}
	\centering
	\includegraphics[width=8.1cm]{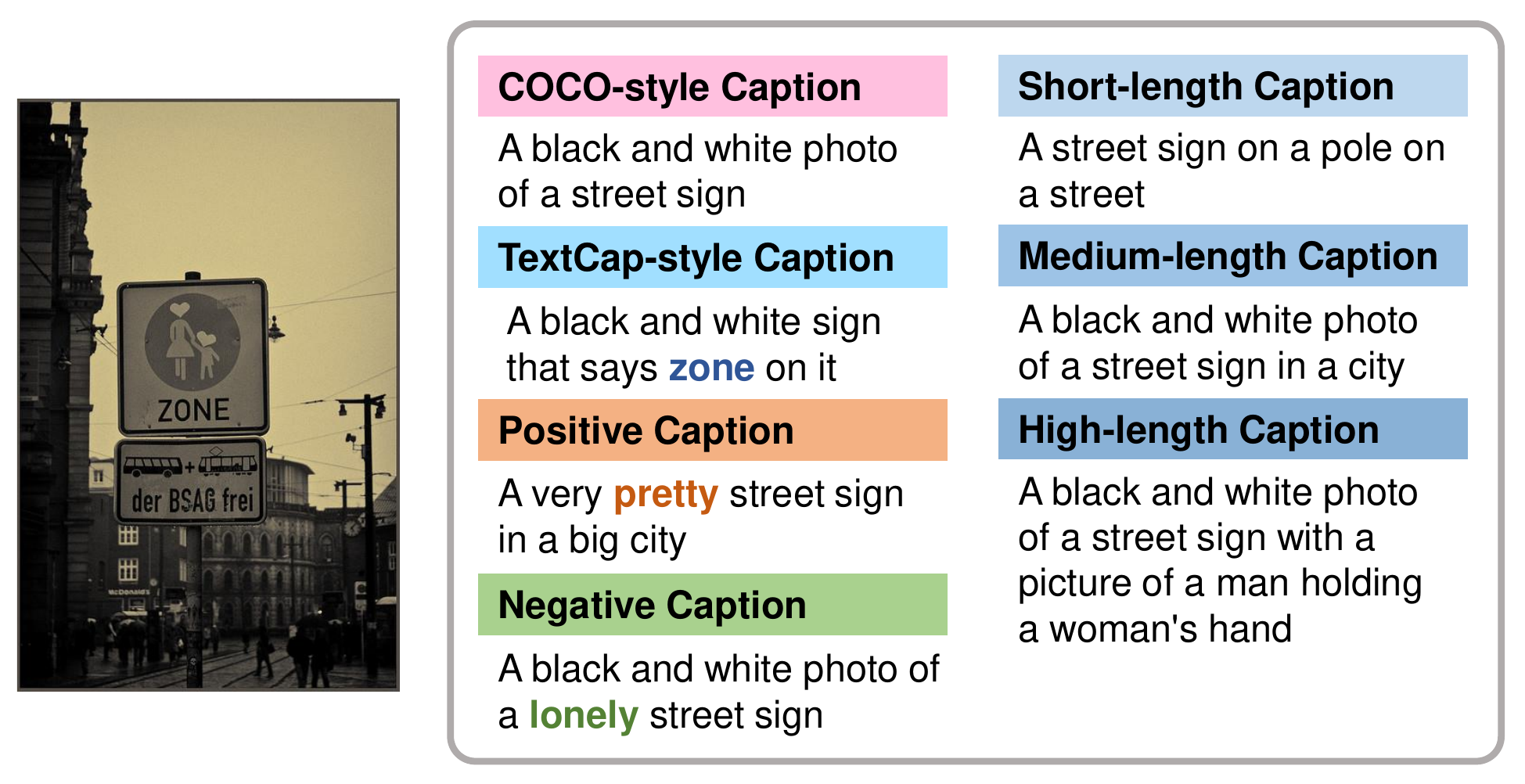}
	\caption{Leveraging a unified model, the proposed method is able to generate diverse captions  such as COCO-style [$ \color[RGB]{255,193,224} \blacksquare$], TexCap-style [$ \color[RGB]{161,224,255} \blacksquare$], Positive [$ \color[RGB]{244,177,131} \blacksquare$], Negative [$ \color[RGB]{169,209,142} \blacksquare$], and different caption lengths including Short-length [$ \color[RGB]{189,215,238} \blacksquare$], Medium-length [$ \color[RGB]{157,195,230} \blacksquare$], and High-length [$ \color[RGB]{138,178,214} \blacksquare$]. Best view in color. 
	}
	\label{fig:intro}
\end{figure}

Despite the remarkable advances, current image captioning algorithms generally lack the controllable capability to generate desired captions.
In other words, once the captioning model is trained, the caption generation process can hardly be influenced.
Typical cases include the control of caption length and description style.
(1) \emph{Length controllable capability}. Sometimes, a brief description is required to get an overview of the image, while in other circumstances, a detailed caption is preferred to acquire more information. 
This can be roughly reflected by the controllable capability of the caption length, which is a basic demand in practical applications, but has been largely overlooked in existing methods.
(2) \emph{Style controllable capability}. 
An image can be described in quite different views. 
For example, given an image with textual contents (e.g., a poster or sign), some people care about the objects, but some may pay more attention to the textual words.  
Besides, people may generate non-factual captions, e.g., emotional descriptions that contain positive or negative expressions.
It is of vital importance to insert different styles in the captioning model to enhance its expressibility. 
%
%
%
How to simultaneously maintain multiple styles and freely switch among them is an open problem.
Existing captioning approaches typically separately handle each scenario, e.g., train a captioner on the COCO dataset \cite{MSCOCO} and train another model on the TextCaps dataset \cite{textcaps}.
As a result, these captioners are domain-specific, without style controllability.

In this paper, we show that a unified model is able to generate captions with different lengths and styles.
As shown in Figure~\ref{fig:intro}, our approach describes an image semantically accurately in diverse views.
This captioning controllable capability is achieved by designing prompts within the cross-modal language model.
After large-scale pre-training, the image captioner has already gained the ability to generate diverse captions, but is largely overwhelmed in the downstream fine-tuning, e.g., on a certain stylized dataset such as COCO \cite{MSCOCO}.
In this work, we aim to unveil the potential hidden in the pre-trained model to flexibly switch captioning styles.
Our approach is motivated by the recent advance in prompt learning techniques \cite{prompt_survey} in natural language processing (NLP).
In the proposed framework, prompts serve as the anchor points to gather data from different domains, facilitating the multi-domain joint training. 
%
%
By virtue of prompt engineering, captions with different lengths, different styles, and different emotions can be properly separated within a unified model.
The prompts, together with the image-text pair, jointly serve as the training corpus to optimize the captioning model. 
%
Furthermore, instead of manually designing prompts, we encourage the captioner to automatically learn the prompt embeddings in an end-to-end manner.
This continuous auto-prompt learning searches the suitable prompt representations in the entire word embedding space, which not only avoids the heuristic prompt design but also exhibits superior performance.

In the inference stage, different prompts serve as the prediction hints to guide the caption generation.
By automatically learning multiple prompt embeddings, the proposed approach has the following merits. Our approach (i) is free of manual prompt engineering, which requires domain expertise and careful word tuning; 
(ii) is able to generate diverse stylized captions via a single model, which is infeasible for most existing state-of-the-art captioners such as BLIP \cite{BLIP}, LEMON \cite{LEMON}, and SimVLM \cite{SimVLM}; 
(iii) does not degrade the performance on different domains such as COCO \cite{MSCOCO} and TextCaps \cite{textcaps}, and outperforms the traditional training strategy using a prefixed prompt;
(iv) is simple and general, which is ready to perform on more domains by incorporating other stylized data.

In summary, the contributions of this work are three-fold:
\begin{itemize}	
		\item To our knowledge, we are the first to propose the prompt-based image captioning framework, which provides a simple yet effective manner to control the caption style. 
		\item We validate the manually designed prompts. We further introduce auto-prompt learning to avoid the heuristic prompt design and achieve superior results. 		
		\item Qualitative and quantitative results verify the controllable capability of the proposed framework. Leveraging a unified model, we achieve outstanding performance on several benchmarks including COCO Karpathy set \cite{MSCOCO}, NoCaps \cite{nocaps}, and TextCaps \cite{textcaps}.
\end{itemize}

\section{Related Work}
{\noindent \bf General Image Captioning.}
Image captioning aims to generate a textual description of the image contents \cite{show-and-tell}, which typically contain a visual encoder to extract the image features and a multi-modal fusion model such as LSTM and Transformer for text generation. 
To represent the visual contents, previous methods \cite{AoANet,BUTD,length_control,meshed-memory-captioning,A2_AAAI,GET_AAAI} utilize the Region-of-Interest (RoI) features from object detectors \cite{FasterRCNN}.
Recent captioning algorithms \cite{ViTCap,E2E-VLP,SimVLM} shed light on the grid features for high efficiency and potentially better performance due to end-to-end training.
%
%
As for the cross-modal model, classic captioners \cite{BUTD,AoANet,XLAN,CAAG_AAAI} typically utilize the LSTM, while the recent approaches \cite{OSCAR,VINVL,BLIP,SimVLM,PureT_AAAI,DLCT_AAAI} leverage the attention-based models to fuse vision-language representations and predict the captions.

{\noindent \bf Controllable Image Captioning.}
Despite the impressive progress, fewer efforts have been made to control the caption generation.
Cornia \emph{et al.} \cite{show_control_tell} utilize image regions to generate region-specific captions.
Chen \emph{et al.} \cite{say_as_you_wish} propose the abstract scene graph to represent user intention and control the generated image captions.
Length-controllable captioning approach is proposed in \cite{length_control}, which learns length level embeddings to control the caption length.
Shuster \emph{et al.} \cite{engaging_personality} release an image captioning dataset with personality traits as well as a baseline approach.
Zhang \emph{et al.} \cite{Magic_AAAI} propose a multi-modal relational graph adversarial inference (MAGIC) framework for diverse text caption.
SentiCap \cite{senticap} utilizes a switching recurrent neural network with word-level regularization to generate emotional captions.
Chen \emph{et al.} \cite{factual_emotional} present a style-factual LSTM to generate captions with diverse styles such as humorous and romantic.
However, some of the aforementioned methods \cite{show_control_tell,say_as_you_wish,factual_emotional} rely on additional tools or expensive annotations for supervision. 
In \cite{kobus2016domain_control}, domain/tag embeddings are involved to control the style, and thus the model architecture is tag-related.
Some methods \cite{senticap,factual_emotional} can be regarded as the ensemble framework, which include two groups of parameters for factual and stylized branches, increasing the model complexity.

In this work, we control the image captioning style from a different view, i.e., prompt learning.
The proposed framework merely involves lightweight learnable prompt embeddings while keeping the baseline architecture unchanged, which is conceptually simple and easy to implement.
%

\begin{figure*}
	\centering
	\includegraphics[width=15.9cm]{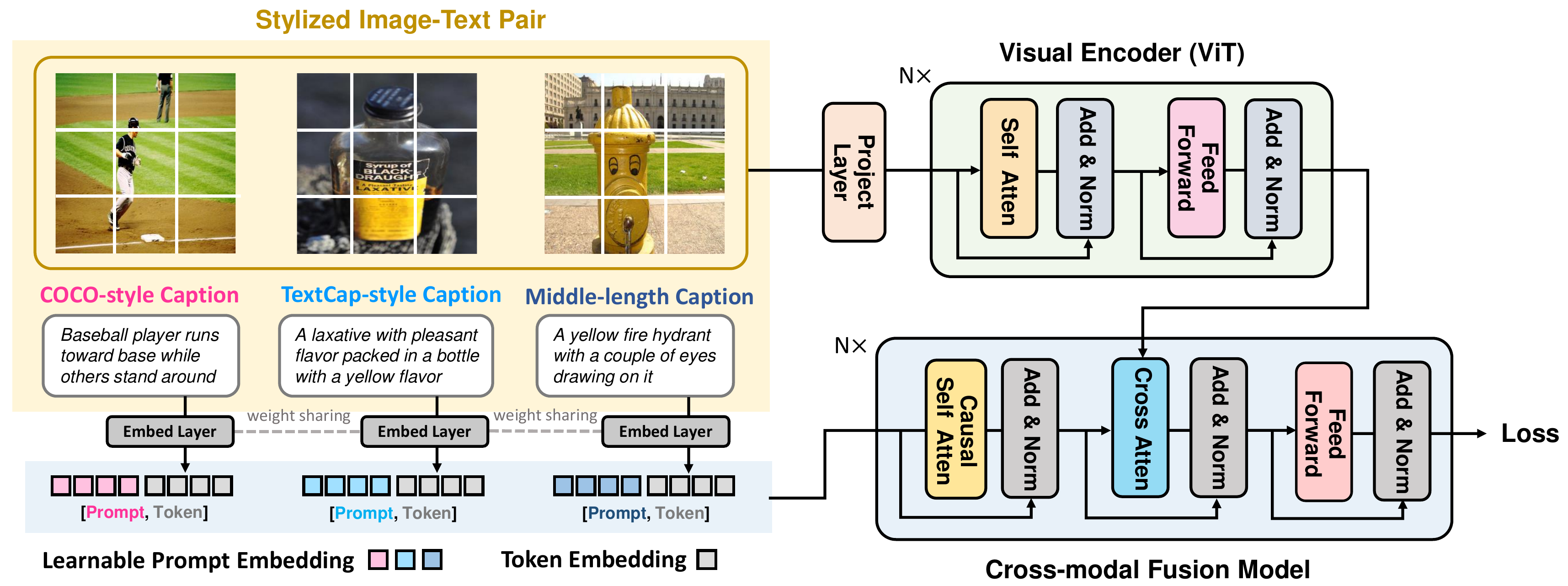}
	\caption{An overview of the proposed prompt-based image captioning framework. Our model optimizes multiple learnable prompt embeddings to absorb stylized data from different domains to jointly train the image captioner. In the inference stage, the model is able to generate diverse captions by feeding different prompts.}
	\label{fig: main framework}\label{fig: main_fig}
\end{figure*}

{\noindent \bf Vision-language Pre-training.}
Vision-language (VL) pre-training is a popular manner to bridge vision and language representations \cite{empirical-study-VL}.
CLIP \cite{CLIP} and ALIGN \cite{ALIGN} use the cross-modal contrastive learning to align the VL representations.
Recent VL pre-training approaches \cite{VLP,UNITER,SOHO} generally adopt the attention mechanism \cite{Transformer} to fuse the VL representations.
After large-scale pre-training on the image-text corpus, these models are further fine-tuned on the downstream datasets to conduct a variety of VL tasks such as image captioning.
SOHO \cite{SOHO} extracts compact image features via a learned visual dictionary and trains the whole framework in an end-to-end manner. 
ALBEF \cite{ALBEF} conducts the cross-modal alignment using contrastive learning technique \cite{CLIP} before representation fusion.
SimVLM \cite{SimVLM} utilizes prefix language modeling for model optimization on the large-scale VL corpus. 
Inspired by previous arts, we also involve VL pre-training to improve the captioning quality.

{\noindent \bf Prompt Learning.}
Prompt learning has gained increasing popularity in natural language processing (NLP) \cite{prompt_survey}.
Prompt learning allows the language model to be pre-trained on the large-scale corpus, and is able to perform downstream tasks by defining a proper prompting function.
Jiang \emph{et al.} \cite{jiang2020can} propose mining-based and paraphrasing-based approaches to automatically generate high-quality prompts.
Shin \emph{et al.} \cite{shin2020autoprompt} search for the proper prompts via a gradient-based approach.
Recently, continuous prompt learning has been explored, which directly optimize prompt vectors in the continuous word embedding space \cite{zhong2021factual,li2021prefix,lester2021power,CoOp}.
It is worth mentioning that prompt learning has been rarely touched in the image captioning.
Different from the traditional usage of prompt learning that aims to elicit knowledge for higher performance, we focus on the controllable capability of the captioning algorithm.
In the proposed framework, except for the superior performance, the more attractive characteristic is that we can freely switch diverse styles via prompting, which greatly enhances the controllability and expressibility of the image captioner.
%


\section{Approach}

In this section, we introduce the method details of the proposed controllable image captioner. 
First, in Section~\ref{method:revisit}, we revisit autoregressive image captioning, which serves as the baseline of our approach.
Then, in Section~\ref{method:prompt}, we elaborate the manual prompt engineering for image captioning.
Finally, we exhibit how to optimize the learnable prompts in Section~\ref{method:auto-prompt} and the inference details in Section~\ref{method:inference}.

\subsection{Revisiting Autoregressive Image Captioning}\label{method:revisit}

In our method, we adopt the unidirectional language modeling (LM) based image captioning framework as the baseline.
Such a framework typically utilizes a transformer block to fuse the image $\boldsymbol{v}$ and text sequence $\boldsymbol{x} = \{x_1, x_2, \cdots, x_n\}$.
The token $x_t$ is generated in an autoregressive manner based on the previous tokens ${\boldsymbol x}_{<t}$.  
The training objective of the cross-modal LM loss is as follows:
\begin{equation}\label{lm loss}
	{\cal L}_{\text{LM}} = - {\mathbb E}_{ (\boldsymbol{v}, \boldsymbol{x}) \in {\cal D}} \Big[ \sum_{t} \text{log}P\left( x_t| g({\boldsymbol v}), f({\boldsymbol x}_{<t})\right) \Big],
\end{equation}
where $g(\cdot)$ denotes the visual encoder, $f(\cdot)$ represents the word embedding layer, $P(\cdot|\cdot)$ can be regarded as the cross-modal fusion model (e.g., transformer decoder in Figure~\ref{fig: main_fig}), which receives the visual features $g({\boldsymbol v})$ and previous token embeddings $f({\boldsymbol x}_{<t})$ to predict the next word token $x_t$.

During inference, the autoregressive models take a special token \texttt{[BOS]} as input to predict
the first token $x_1$, then $x_1$ is fed into the model to obtain the next token $x_2$.
This autoregressive prediction process is continued until the special token \texttt{[EOS]} is predicted.

\subsection{Prompt-based Image Captioning}\label{method:prompt}

{\flushleft \bf Model Pre-training.} Following previous works \cite{VINVL,LEMON,BLIP,SimVLM}, we also adopt the large-scale pre-training on the noisy image-text corpus to improve the downstream captioning task.
Besides the language modeling (LM) loss, we also adopt the image-text contrastive loss \cite{CLIP,ALIGN} and image-text matching loss \cite{UNITER,ALBEF,BLIP} to jointly optimize the visual encoder and cross-modal fusion model, as follows:
\begin{equation}\label{caption loss}
	{\cal L}_{\text{Pre-train}} = {\cal L}_{\text{Contrast}} + {\cal L}_{\text{Match}} + {\cal L}_{\text{LM}}.
\end{equation}
The contrastive loss measures the similarity of the image-text pairs via a light fusion manner such as dot-product, while the matching loss measures the image-text similarity via a heavy fusion manner such as cross-attention.
It has been widely recognized that both of them can facilitate cross-modal alignment \cite{BLIP}.
Therefore, although we focus on the image captioning, we additionally include the ${\cal L}_{\text{Contrast}} $ and ${\cal L}_{\text{Match}} $ in the pre-training stage.
As for more details, please refer to BLIP \cite{BLIP}.
%

{\flushleft \bf Prompt Engineering.}
After pre-training, the model already acquires zero-shot captioning capability thanks to the language modeling loss ${\cal L}_{\text{LM}}$.
%
%
Therefore, previous LM-based image captioners such as SimVLM \cite{SimVLM} and BLIP \cite{BLIP} leverage a pre-defined prompt such as ``\texttt{a picture of}'' or ``\texttt{a photo of}'' to facilitate the image captioning.
In this work, we aim to unveil the model potential of generating diverse captions via prompting.

In contrast to single prompt engineering, in the fine-tuning stage, we design multiple prompts as the anchors to distinguish the training data from different domains.
In this way, different stylized captions do not disturb their counterparts and together contribute to a stronger model.
%
%
The manually designed prompts are illustrated in Table~\ref{table:manual_prompts}.
(i) For the cross-domain scenario, e.g., evaluating a model on both COCO \cite{MSCOCO} and TextCaps \cite{textcaps}, it is straightforward to assign different prompts for these datasets to learn domain-specific descriptions.
(ii) As for the caption length control, we divide the image captions from COCO and TextCaps into three levels depending on the caption length.
Captions whose length is in the range $[0,10)$, $[10,16)$, and $[16,+\infty)$ are divided.
Each of these subsets is assigned with a specific prompt, as shown in Table~\ref{table:manual_prompts}.   
(iii) Finally, current image captions are typically factual.
Nevertheless, each image in the COCO dataset is labeled by five annotators, inevitably containing emotional descriptions.
To this end, we collect the positive and negative captions in the COCO dataset to form the 
non-factual subsets, which contain the pre-defined positive words such as ``\texttt{great}, \texttt{nice}, \texttt{cute}'' and negative words such as ``\texttt{ugly}, \texttt{terrible}, \texttt{disgusting}''.
Despite these non-factual captions being rare, our method still learns satisfying styles using limited samples, justifying the few-shot learning ability of prompt engineering.
The entire positive and negative words, and other potentially effective manual prompts are presented in the \emph{supplementary material}.

{\flushleft \bf Model Fine-tuning.} 
In our framework, multiple training sets are mixed together to train a unified model.
%
Compared to Eq.~(\ref{lm loss}), we predict token $x_t$ based on the visual features $g({\boldsymbol v})$, prompt token embeddings $f({\boldsymbol p})$, and previous token embeddings $f({\boldsymbol x}_{<t})$.
Different stylized data is assigned with a specific prompt as illustrated in Table~\ref{table:manual_prompts}.
During training, we prepend these hand-crafted prompts to caption tokens as the textual description of the image.
We assemble different stylized datasets to jointly train the captioning model using a prompt-based LM loss $ {\cal L}_{\text{ProLM}} $ as follows:
%
\begin{equation}\label{caption loss}
	\scriptsize
	{\cal L}_{\text{ProLM}} = - \sum_i \left[ {\mathbb E}_{ \left(\boldsymbol{v}, \boldsymbol{p}_i, \boldsymbol{x} \right) \in {\cal D}_i} \Big[ \sum_{t} \text{log}P \left( x_t| g({\boldsymbol v}), f({\boldsymbol p}_i), f({\boldsymbol x}_{<t}) \right) \Big] \right],
\end{equation}
where $\boldsymbol{p}_i$ denotes the manual prompt for $i$-th dataset ${\cal D}_i$.
Note that the prompt tokens ${\boldsymbol p}_i$ and caption tokens ${\boldsymbol x}_{<t} $ share the same embedding mapping layer $f(\cdot)$.
In this framework, we keep the baseline model architecture unchanged without additional learnable blocks, which is parameter-efficient.

\setlength{\tabcolsep}{2pt}
\begin{table}
	\scriptsize
	\begin{center}
		\begin{tabular*}{8.2 cm} {@{\extracolsep{\fill}}l|c}
			\hline
			~ Caption Style  &Manual Prompt $\boldsymbol{p}$   \\
			\hline
			~   COCO-style &\texttt{a normal picture that shows}      \\
			~   TextCap-style &\texttt{a textual picture that shows}  \\ 
			~   Short-length & \texttt{a picture with a short caption that shows}   \\
			~   Medium-length & \texttt{a picture with a medium caption that shows} \\
			~   High-length  &\texttt{a picture with a long caption that shows}  \\
			~   Positive &\texttt{a positive picture that shows}\\ 
			~   Negative &\texttt{a negative picture that shows} \\ 
			\hline
		\end{tabular*}
		\caption{Illustration of the manual prompts.} \label{table:manual_prompts}
	\end{center}
\end{table}

\subsection{Auto-prompt Learning}\label{method:auto-prompt}

To avoid the laborious manual prompt engineering in Section~\ref{method:prompt}, we further encourage the network to automatically learn the prompts in an end-to-end manner, as shown in Figure~\ref{fig: main_fig}.
Given a sequence of the manual prompt tokens such as ``\texttt{a textual picture that shows}'', the model first maps each token to a unique numeric ID using WordPiece technique.
Then, for a BERT-base model, the token IDs are projected to 768-dim word embeddings via the token embedding layer $f(\cdot)$ as the input of the vision-language fusion model, i.e., $f({\boldsymbol p}) \in \mathbb{R}^{N \times 768}$, where $N$ represents the prompt length.
%
Instead of the manual prompt engineering, we propose to learn the caption prompt embeddings ${\bf P}$ as follows:
\begin{equation}\label{prompt_embedding}
    \small
	{\bf P} = [\mathrm{P}]_1[\mathrm{P}]_2\cdots[\mathrm{P}]_N,
\end{equation}
where each embedding vector $[\mathrm{P}]_k~(k\in{1,\cdots,N})$ has the same dimension as the word embedding. In other words, ${\bf P} \in \mathbb{R}^{N \times 768}$ serves as an alternative of the manual prompt embedding $f({\boldsymbol p})$.
In the training stage, prompt embeddings $\bf P$ are jointly optimized with the captioning network as follows:
\begin{equation}\label{auto-prompt loss}
	\small
	{\cal L}_{\text{AutoProLM}} = - \sum_i \left[ {\mathbb E}_{ \left(\boldsymbol{v}, \boldsymbol{x} \right) \in {\cal D}_i} \Big[ \sum_{t} \text{log}P \left( x_t| g({\boldsymbol v}), {\bf P}_i, f({\boldsymbol x}_{<t}) \right) \Big] \right].
\end{equation}

The proposed framework learns specific prompt embeddings ${\bf P}_i$ for each domain-specific dataset ${\cal D}_i$.
During the end-to-end training, the gradients can be effectively back-propagated to optimize the prompt embeddings. 
To this end, the captioner is able to fully explore the suitable prompt representations in the continuous word embedding space.

\subsection{Prompt-based Inference}\label{method:inference}

After prompt learning, our model is able to generate diverse captions using different prompts.
In the manual prompt framework, after encoding the special token \texttt{[BOS]}, we sequentially embed the prompt tokens via $f({\boldsymbol p})$ and feed them to the language model to generate the caption in an autoregressive manner.
In the auto-prompt framework, we directly concatenate the token embedding of \texttt{[BOS]} and learned prompt embeddings $\bf P$ as the input of the language model. 
By switching different prompts, the proposed captioner is able to generate a certain stylized caption.

\section{Experiment}

\subsection{Datasets and Metrics}\label{section:dataset and metric}

{\bf \noindent Pre-training Data.} In the experiments, following our baseline approach \cite{BLIP}, we collect the image-text pairs from Visual Genome \cite{VG}, COCO \cite{MSCOCO}, SBU Captions \cite{SBU}, Conceptual Captions \cite{CC3M}, Conceptual 12M \cite{CC12M}, and a filtered version of LAION (115M images) \cite{LAION} to form the pre-training data.
Following BLIP \cite{BLIP}, these data are filtered by a large model to form the high-quality bootstrapped dataset.
In total, the pre-training corpus consists of about 129 million images. 
%

{\bf \noindent Evaluation Datasets and Metrics.} We evaluate the proposed method on the COCO caption dataset \cite{MSCOCO} of Karpathy split \cite{karpathy}, NoCaps \cite{nocaps}, and TextCaps \cite{textcaps}.
%
%
%
To evaluate the quality of the generated captions, we use standard metrics in the image captioning task, including BLEU@4 \cite{BLEU}, METEOR \cite{METEOR}, CIDEr \cite{CIDER}, and SPICE \cite{SPICE}. 
In the inference stage, beam search (beam size = $3$) is adopted in all experiments.
More details and visualization results can be found in the supplementary material.

\subsection{Implementation Details}\label{section: implementation details}

Our model is implemented in Python with PyTorch.
In the pre-training stage, the model is trained on 32 V100 GPUs. 
The image encoder is initialized from ViT-B/16 pre-trained on the ImageNet \cite{ViT}, and the text encoder is initialized from BERT-base \cite{BERT}.
We pre-train the whole model for 32 epochs using a batch size of 2880.
We use AdamW optimizer \cite{Adam} with a weight decay of 0.05. 
The learning rate is warmed-up to $3\times10^{-4}$ and decayed linearly with a rate of 0.85. 
We take random image crops of resolution $224\times224$ during pre-training.

In the fine-tuning stage, we train the model using a small learning rate of $1\times10^{-5}$ and linearly decay it.
The model is fine-tuned for $5$ epochs.
Following previous works \cite{SimVLM}, the image resolution is increased to $384 \times 384$ during fine-tuning.
%
%
As for the prompt embedding ${\bf P} \in \mathbb{R}^{N \times 768}$, we randomly initialize it and set $N=16$.
We optimize our algorithm using standard cross-entropy loss \emph{without} reinforcement learning.
The proposed \underline{Con}trollable \underline{Cap}tioner is denoted as {ConCap}. 
%
%

\setlength{\tabcolsep}{2pt}
\begin{table}[h]
	\scriptsize
	\begin{center}
		\begin{tabular*}{8.3 cm} {@{\extracolsep{\fill}}l|cc|cc}
			\hline
			\multirow{2}{*}{Configuration}  & \multicolumn{2}{c|}{COCO Test} & \multicolumn{2}{c}{TextCaps Val} \\
			 &B@4 &C &B@4 &C \\
			\hline
			 {\scriptsize \bf w/o Fine-tuning (Frozen Model)}  &  &  &  & \\
			 \ding{172} w/o Prompt &{ 33.9}  &{ 106.6}  &{ 18.6} &{ 48.6} \\
			 \ding{173} Manual Prompt (i.e., \texttt{a picture of}) &23.7 &83.8  &14.5 &38.4\\
			 \ding{174} Learned Prompt ($N=16$) &{\bf 38.3} &{\bf 125.1}  &{\bf 20.7} &{\bf 56.7}\\
			\hline
			 {\scriptsize \bf Multi-dataset Individual Training}  &  &  & & \\
			 \ding{175} w/o Prompt &39.1  &132.4  &{ 30.4} &{ 113.4} \\
			 \ding{176} Manual Prompt (i.e., \texttt{a picture of}) &{ 39.4} &{ 132.6}  &30.1  &111.2 \\
			 \ding{177} Learned Prompt ($N=16$) &{\bf 40.5} &{\bf 133.5}  &{\bf 31.2} &{\bf 115.9}\\
			\hline
			 {\scriptsize \bf Multi-dataset Joint Training}  &  &  & & \\
			 \ding{178} w/o Prompt  &39.2  &131.9  &30.1  &111.6 \\
			 \ding{179} Shared Manual Prompt (i.e., \texttt{a picture of})    &39.3  &132.2  &30.0  &110.4 \\
			 \ding{180} Multi-prompt (Manual Prompts in Table~\ref{table:manual_prompts})    &39.6  &132.8  &30.7 &113.5 \\
			 \ding{181} Multi-prompt (Auto Learning, $N=16$)  &{\bf 40.5}  &{\bf 133.7}  &{\bf 31.3}  &{\bf 116.7} \\
			\hline
		\end{tabular*}
		\caption{Ablation comparisons on the COCO Karpathy test split \cite{MSCOCO} and TextCaps validation set \cite{textcaps}, where B@4 and C denote BLEU@4 and CIDEr scores, respectively.} \label{table: prompt ablation}
	\end{center}
\end{table}

\subsection{Ablation Study}

{\noindent \bf Manual Prompt \emph{v.s.} w/o Prompt.} 
%
%
Previous works such as BLIP utilize a pre-defined prompt ``\texttt{a picture of}'' to facilitate the caption generation.
However, as shown in Table~\ref{table: prompt ablation}, in the zero-shot evaluation without model fine-tuning (\ding{172} and \ding{173}), an empty prompt is even more effective. 
After downstream model fine-tuning (\ding{175} and \ding{176}), we observe that this hand-crafted prompt is beneficial to COCO dataset but harmful to TextCaps. 
These results show that the heuristic prompt is not always a good choice, which potentially requires the laborious manual design for different datasets.

%

{\noindent \bf Effectiveness of Learned Prompt.} 
For a \emph{frozen} image captioner, we only optimize the prompt embeddings in setting \ding{174} in Table~\ref{table: prompt ablation}.
The results show that learned prompt embeddings greatly unveil the potential of a pre-trained model with a good zero-shot performance of 125.1 CIDEr on COCO.
After joint training of prompt embeddings and captioning model, the performance of the learned prompt is still superior to the manual prompt (\ding{177} \emph{v.s.} \ding{176}).

{\bf \noindent Multi-dataset Individual Training \emph{v.s.} Joint Training.}
Previous works typically train the model individually on different datasets.
In setting \ding{176} in Table~\ref{table: prompt ablation}, we separately fine-tune the image captioner on COCO and TextCaps.
In setting \ding{179} in Table~\ref{table: prompt ablation}, we merge the datasets of COCO and TextCaps, and use a \emph{shared} prompt ``\texttt{a picture of}'' for both datasets.
By analyzing the results of \ding{176} and \ding{179}, we can observe that simply combining two diverse datasets with different styles will degrade the performance.
This is consistent with common sense that data from diverse domains will challenge the model training.

{\noindent \bf Single Prompt \emph{v.s.} Multi-prompt.} In setting \ding{180} of Table~\ref{table: prompt ablation}, we still combine the COCO and TextCaps to jointly train a unified captioner, but separate multi-domain data using different prompts.
%
It is interesting that ``multi-prompt for joint training'' (\ding{180}) not only outperforms the ``single prompt for joint training'' (\ding{179}), but also surpasses ``single prompt for individual training'' (\ding{176}), indicating that multiple (even manually designed) prompts can effectively separate the data from different domains.
Furthermore, the most promising characteristic of ``multi-prompt'' is that we can control the caption style by feeding different prompts, which is infeasible for the ``single prompt'' setting.
%
%
%
Finally, we encourage the model to jointly optimize multiple learnable prompt embeddings in an end-to-end manner (\ding{181}), which achieves the best results.

%

%
%
%

{\noindent \bf Auto-prompt Length.}
We further validate the influence of prompt embedding length $N$.
In Table~\ref{table:prompt length}, the prompt embeddings of different lengths are randomly initialized.
We test different lengths of $N=1,4,8,16,24$ and observe that increasing the prompt embedding length $N$ can consistently improve the performance.
In our experiments, We choose $N=16$ as it already yields saturated results.

\begin{table}
	\scriptsize
	\centering
	
	\begin{tabular*}{8.5 cm} {@{\extracolsep{\fill}}l|cc|cc|cc|cc|cc}
		\hline
		  \multirow{2}{*}{}  &\multicolumn{2}{c|}{$N=1$} &\multicolumn{2}{c|}{$N=4$} &\multicolumn{2}{c|}{$N=8$} &\multicolumn{2}{c|}{$N=16$} &\multicolumn{2}{c}{$N=24$} \\
		   &B@4 &C &B@4 &C &B@4 &C &B@4 &C &B@4 &C  \\
		\hline
		  TextCaps Val &30.7  &114.5 &30.8 &115.4 &31.1 &115.4 &31.3 &{\bf 116.7} &{\bf 31.5} &115.9 \\
		\hline
	\end{tabular*}
	\caption{Ablation of the prompt embedding length $N$ on the TextCaps validation set \cite{textcaps}.} \label{table:prompt length}
\end{table}

{\noindent \bf Evaluation of Different Prompts on COCO}.
Finally, we evaluate the performance of different automatically learned prompts on the COCO Karpathy test split. The results are shown in Table~\ref{table: different prompts on COCO}.
There is no doubt that ``COCO-style Prompt'' overall performs best, which leverages the entire training set of COCO for model fine-tuning.
``TextCap-style Prompt'' exhibits poor results on the COCO dataset, which justifies the domain gap between COCO and TextCaps datasets. 
Finally, it is interesting that CIDEr metric \cite{CIDER} prefers a short caption and ``Short-length Prompt'' is even comparable to the best ``COCO-style Prompt'' in CIDEr metric, while SPICE metric \cite{SPICE} prefers a longer caption and ``High-length Prompt'' clearly outperforms the strong ``COCO-style Prompt'' in SPICE.
%
Visualization results of different prompts are shown in the next Section~\ref{sec: visualization}.

\begin{table}
	\scriptsize
	\centering
	\begin{tabular*}{6.5 cm} {@{\extracolsep{\fill}}l|cccc}
		\hline
		 Prompt Style  &B@4 &M &C &S  \\
		\hline
		 Short-length Prompt &39.9 &30.2 &132.3 &23.0  \\
		 Medium-length Prompt  &35.1 &30.9 &122.9 &23.9  \\
		 High-length Prompt  &26.9 &30.7 &71.6 &{\bf 25.0}  \\
		 Positive Prompt  &27.0 &25.8 &97.6 &20.7 \\
		 Negative Prompt   &37.0 &29.3 &121.5 &22.9  \\
		 TextCap-style Prompt &22.1 &25.9 &66.0 &20.5 \\
		 COCO-style Prompt &{\bf 40.5} &{\bf 30.9} &{\bf 133.7} & 23.8 \\
		\hline
	\end{tabular*}
	\caption{Performance comparisons of different prompts on the COCO Karpathy test split \cite{MSCOCO}, where B@4, M, C, S denote BLEU@4, METEOR, CIDEr, and SPICE.} \label{table: different prompts on COCO}
\end{table}

\subsection{Qualitative Evaluation}\label{sec: visualization}

\begin{figure}[t]
	\centering
	\includegraphics[width=8.5cm]{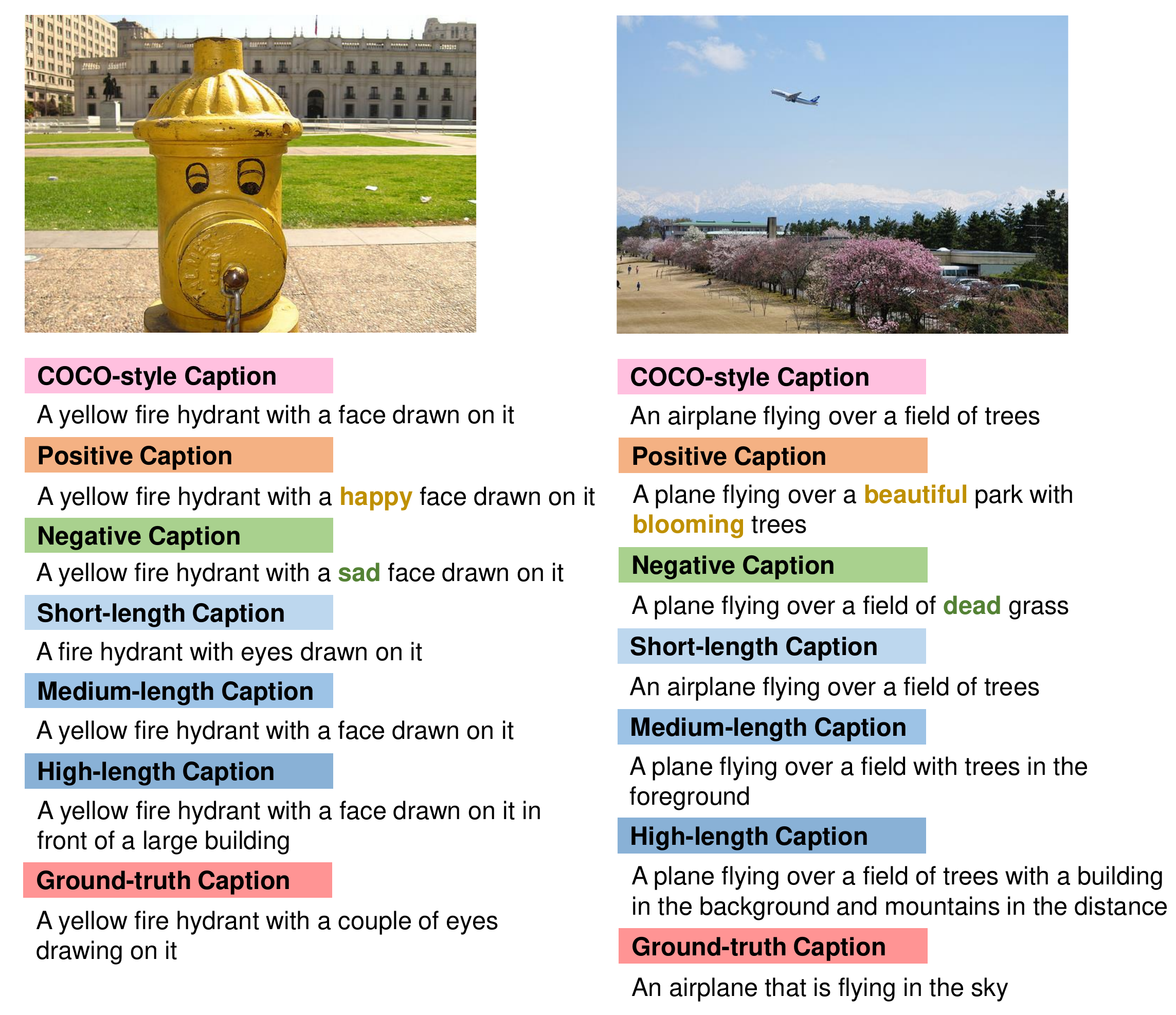}
	\caption{Image captioning examples from COCO \cite{karpathy} with different styles including COCO-style [$ \color[RGB]{255,193,224} \blacksquare$], Positive [$ \color[RGB]{244,177,131} \blacksquare$], Negative [$ \color[RGB]{169,209,142} \blacksquare$], Short-length [$ \color[RGB]{189,215,238} \blacksquare$], Medium-length [$ \color[RGB]{157,195,230} \blacksquare$] and High-length [$ \color[RGB]{138,178,214} \blacksquare$].}
	\label{fig:coco_example}
\end{figure}

\begin{figure}[h]
	\centering
	\includegraphics[width=8.5cm]{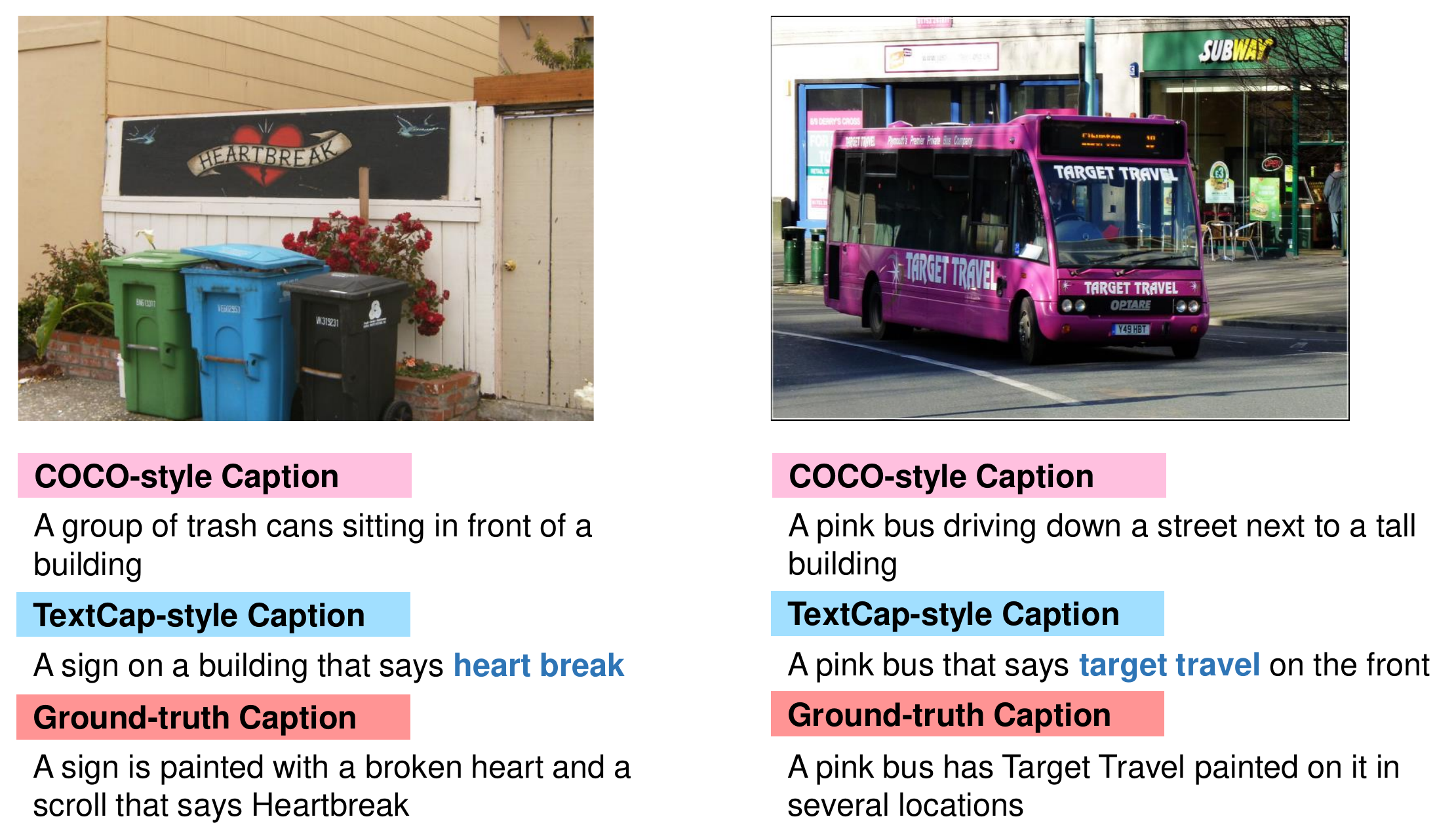}
	\caption{Image captioning examples from TextCaps dataset \cite{textcaps} with different styles including COCO-style [$ \color[RGB]{255,193,224} \blacksquare$] and TexCap-style [$ \color[RGB]{161,224,255} \blacksquare$]. Best view in zoom in.}
	\label{fig:textcap_example}
\end{figure}

{\noindent \bf Results on COCO \cite{MSCOCO}.} In Figure~\ref{fig:coco_example}, we exhibit the captioning results on the COCO dataset.
%
%
By feeding different prompts, our ConCap method is able to generate diverse captions including COCO-style [$ \color[RGB]{255,193,224} \blacksquare$], Positive [$ \color[RGB]{244,177,131} \blacksquare$], Negative [$ \color[RGB]{169,209,142} \blacksquare$], Short-length [$ \color[RGB]{189,215,238} \blacksquare$], Medium-length [$ \color[RGB]{157,195,230} \blacksquare$], and High-length [$ \color[RGB]{138,178,214} \blacksquare$].
%
%
Besides, we observe that the percentage of emotional captions is only about 2\% of the entire COCO dataset.
The proposed ConCap merely utilizes limited positive or negative captions in COCO to learn such styles.
This is consistent with the observation that prompt learning is suitable for few-shot domain transfer \cite{prompt_survey}.
%
%
As shown in Figure~\ref{fig:coco_example}, our ConCap is able to briefly describe an image or in a more detailed manner.
The high-length captions [$ \color[RGB]{138,178,214} \blacksquare$] produced by our ConCap are much longer than the ground-truth captions and yield additional meaningful semantics, e.g., ``\texttt{a large building}'' in the first image and ``\texttt{mountains in the distance}'' in the second image.  
Furthermore, our approach generates the positive words such as ``\texttt{happy}, \texttt{beautiful}'' [$ \color[RGB]{244,177,131} \blacksquare$] or the negative words such as ``\texttt{sad}, \texttt{dead}'' [$ \color[RGB]{169,209,142} \blacksquare$] to describe the same image in opposite personality traits.
Since the COCO-caption dataset rarely contains the image with OCR contents, we showcase the results of ``TextCap-style Prompt'' on the TextCaps dataset in Figure~\ref{fig:textcap_example}.

\setlength{\tabcolsep}{2pt}
\begin{table*}[h]
	\scriptsize
	\begin{center}
		\begin{tabular*}{15.8 cm} {@{\extracolsep{\fill}}l|c|cccc|cccccccc}
			\hline
			 \multirow{3}{*}{Method} &\multirow{3}{*}{Pre-training Data} & \multicolumn{4}{c|}{COCO Caption}  & \multicolumn{8}{c}{NoCaps Validation} \\
			
			 & & \multicolumn{4}{c|}{Karpathy Test}  & \multicolumn{2}{c}{In-domain} & \multicolumn{2}{c}{Near-domain} & \multicolumn{2}{c}{Out-domain}  & \multicolumn{2}{c}{Overall} \\
			
			  & &B@4 &M &C &S &C &S &C &S &C &S &C &S \\
			\hline
			
			 BUTD \cite{BUTD}   &N/A &36.2 &27.0 &113.5 &20.3 &80.0 &12.0 &73.6 &11.3 &66.4 &9.7 &73.1 &11.1\\
			 AoANet \cite{AoANet}   &N/A &37.2 &28.4 &119.8 &21.3  &- &- &- &- &- &- &- &-\\
			 X-LAN \cite{XLAN}   &N/A &38.2 &28.8 &122.0 &21.9  &- &- &- &- &- &- &- &-\\
			\hline
			
			 $\text{Oscar}_\text{base}$ \cite{OSCAR}  &7M &36.5 &30.3 &123.7 &23.1  &83.4 &12.0 &81.6 &12.0 &77.6 &10.6 &81.1 &11.7 \\
			
			 $\text{ViTCAP}$ \cite{ViTCap} &10M  &36.3 &29.3 &125.2 &22.6 &98.7 &13.3 &92.3 &13.3 &95.4 &12.7 &93.8 &13.0 \\
			
			 $\text{VinVL}_\text{base}$ \cite{VINVL} &9M  &38.2 &30.3 &129.3 &23.6  &103.1 &14.2 &96.1 &13.8 &88.3 &12.1 &95.5 &13.5 \\
			
			 $\text{LEMON}_\text{base}$ \cite{LEMON} &200M &40.3 &30.2 &133.3 &23.3 &107.7 & 14.7 & 106.2 & 14.3 & 107.9 & 13.1 & 106.8 & 14.1 \\
			
			 $\text{BLIP}_\text{base}$ \cite{BLIP} &129M &39.7 &- &133.3 &23.3 &111.8 &14.9 &{\bf 108.6} &{\bf 14.8} &111.5 &14.2 &109.6 &14.7 \\
			
			 \color{gray} $ \text{SimVLM}_\text{base}$ \cite{SimVLM} &\color{gray} 1.8B &\color{gray} 39.0 & \color{gray} 32.9 &\color{gray} 134.8 &\color{gray} 24.0 &\color{gray} - &\color{gray} - &\color{gray} - &\color{gray} - &\color{gray} - &\color{gray} - &\color{gray} 94.8 &\color{gray} 13.1 \\
			
			 {\bf ConCap (Ours)} &129M &{\bf 40.5} &{\bf 30.9} &{\bf 133.7} &{\bf 23.8} &{\bf 113.4} &{\bf 14.9} &108.4 &14.6 &{\bf 113.2} &{\bf 14.4} &{\bf 110.2} &{\bf 14.8} \\
			\hline
		\end{tabular*}
		\caption{Performance comparisons on the COCO Karpathy test split \cite{MSCOCO} and NoCaps validation split \cite{nocaps}, where B@4, M, C, S denote BLEU@4, METEOR, CIDEr, and SPICE scores. For a fair comparison, all the methods only adopt the standard cross-entropy without CIDEr optimization.} \label{table:SOTA on coco nocaps}
	\end{center}
\end{table*}

{\noindent \bf Results on TextCaps \cite{textcaps}.}
Figure~\ref{fig:textcap_example} exhibits the results on the TextCaps dataset, where we show the COCO-style [$ \color[RGB]{255,193,224} \blacksquare$] and TextCap-style [$ \color[RGB]{161,224,255} \blacksquare$] captions for style comparison.
Different styles focus on different aspects of the image.
For example, in the first image, the TextCap-style caption as well as the ground-truth annotation aim to describe the words in the sign (e.g., ``\texttt{heart break}'') while ignoring the objects such as ``\texttt{trash cans}''.
In contrast, the COCO-style pays more attention to the objects and environment, e.g., ``\texttt{tall building}'' in the second image.

\subsection{Quantitative Evaluation}

{\bf \noindent COCO \cite{MSCOCO}.} In Table~\ref{table:SOTA on coco nocaps}, we present the performance of state-of-the-art captioning methods on the COCO-caption Karpathy test split \cite{karpathy}.
Compared with the recent LEMON \cite{LEMON} that leverages more pre-training data, our method achieves superior performance.
BLIP \cite{BLIP} can be regarded as the baseline of our approach. Compared with BLIP, our ConCap outperforms it on all metrics.
More importantly, the proposed ConCap is able to simultaneously handle other domains and generate captions with different lengths and styles for each image, which is infeasible for BLIP. 
The recent SimVLM approach \cite{SimVLM} leverages a large-scale pre-training corpus including 1.8 billion image-text pairs, which is $10\times$ larger than ours.
Besides, SimVLM combines the ResNet \cite{ResNet} and ViT \cite{ViT} models as the visual extractor, which is stronger than our pure ViT structure.
%

{\bf \noindent NoCaps \cite{nocaps}.} 
NoCaps dataset covers more than 600 object categories and nearly 2/3 of them are unseen from the training set in COCO.
The images in NoCaps are categorized into in-domain, near-domain, and out-of-domain based on whether these images are seen in the COCO training set.
%
%
On this benchmark, we evaluate our ConCap using the ``COCO-style'' prompt.
As shown in Table~\ref{table:SOTA on coco nocaps}, the proposed ConCap outperforms all existing methods in terms of the overall performance, which verifies the generalizability of our method.

{\bf \noindent TextCaps \cite{textcaps}.} TextCaps is a recently proposed dataset containing 28K images and 145K captions, which is more challenging than COCO due to the existence of complex textual words.
We compare the proposed method with the classic captioner such as AoANet \cite{AoANet} and the recent state-of-the-art methods including MMA-SR \cite{MMASR}, CNMT \cite{CNMT}, and TAP \cite{TAP}.

\begin{table}[h]
	\scriptsize
	\centering
	\begin{tabular*}{8.0 cm} {@{\extracolsep{\fill}}l|c|cc|cc}
		\hline
		  \multirow{2}{*}{Method} &OCR &\multicolumn{2}{c|}{Validation}  &\multicolumn{2}{c}{Test} \\
		   &Input &B@4 &C &B@4 &C  \\
		\hline
		   BUTD \cite{BUTD} &\ding{55} &20.1 &41.9 &14.9 &33.8 \\
		   AoANet \cite{AoANet} &\ding{55} &20.4  &42.7 &15.9 &34.6\\
		   M4C \cite{M4C} &\ding{51} &23.3 &89.6  &18.9 &81.0 \\
		   MMA-SR \cite{MMASR} &\ding{51} &24.6 &98.0 &19.8 &88.0 \\ 
		   CNMT \cite{CNMT} &\ding{51} &24.8  &101.7  &20.0 &93.0 \\
		   TAP \cite{TAP} &\ding{51} &25.8 &109.2 &21.9 &103.2 \\ 
		   {\bf ConCap (Ours)} &\ding{55} &{\bf 31.3} &{\bf 116.7} &{\bf 27.4} &{\bf 105.6} \\ 
		\hline
	\end{tabular*}
    \caption{Comparsion results on the TextCaps validation set and test set \cite{textcaps}, where B@4 and C denote BLEU@4 and CIDEr scores, respectively.} \label{table:textcap_results}
\end{table}

The comparison results are shown in Table~\ref{table:textcap_results}.
Our approach significantly outperforms the classic methods without pre-training such as AoANet \cite{AoANet}.
To the best of our knowledge, TAP \cite{TAP} represents the current performance leader on the TextCaps dataset.
TAP approach collects high-quality OCR-based image-text pre-training data, and performs the text-aware pre-training.
Besides, TAP feeds the OCR detection results to the model, while our approach is free of such necessity.
Without knowing the OCR results, our approach still surpasses the current state-of-the-art TAP method by a large margin of 7.5 CIDEr on the validation set.
It is worth noting that our ConCap is not specially designed for TextCaps and is able to perform well on multiple domains including COCO, NoCaps, and TextCaps using a single model.

\section{Conclusion}
In this paper, we propose a conceptually simple yet effective prompt-based image captioning framework, which has been rarely investigated in the captioning community.
By prompt engineering, the proposed approach is able to generate captions with diverse styles.
To further explore the potential of prompt learning, we encourage the network to automatically learn the suitable prompt vectors in the continuous word embedding space.
Extensive qualitative and quantitative experiments verify the effectiveness of the proposed framework.

{	
	\small
	\bibliographystyle{aaai}
	\bibliography{egbib}
}

\clearpage

\appendix

\section{Social Impact and Ethics Statement}\label{section:ethics}
The proposed framework has the following potential positive impacts: 
(1) this work focuses on a meaningful direction of image captioning, i.e., maintaining both style controllability and state-of-the-art performance on prevalent benchmarks;
(2) without bells and whistles, this simple pipeline is general and can be easily combined with existing captioning methods; 
(3) this is the first attempt to absorb the continuous prompt learning idea in the image captioning community, which may inspire future works to explore better formulations in this direction for superior results.

Nevertheless, since our method aims to describe the image from different views, it also potentially raises the risk of privacy invasion (e.g., describing a person in a negative view), which is a common concern in the captioning area.

\section{Method Limitation}

The proposed framework optimizes \emph{learnable} prompts for image captioners.
The merits are three-fold: (1) it avoids laborious manual design; (2) it enhances the model controllable capability and allows the captioner to generate stylized captions using one model; (3) it achieves better results compared with the manual prompt.

Nevertheless, compared with the manual prompt such as ``\texttt{a picture of}'', the main limitation is that the learned prompt representations are difficult to visualize.
Since learned prompts are optimized in the \emph{continuous} word embedding space, they fail to map back to the words in the dictionary.
To this end, we search within the vocabulary for words that are closest to the learned prompt embeddings based on the dot-product similarity.
Table~\ref{table: learned prompts} shows some examples of the approximate learned prompts (token IDs are mapped to tokens via BERT tokenizer).
Although these approximate auto-prompts are difficult to be understood by humans, they are suitable for the image captioner.

Besides, our generated captions also contain some unsatisfactory descriptions.
The failure cases include (1) some high-length captions include redundant expressions; (2) some images fail to generate positive or negative captions. However, we think some failures are also reasonable. For example, forcing a positive image (e.g., a smiling person) to generate negative captions is somewhat difficult.

\section{More Implementation Details}

{\bf \noindent Pre-training Framework.} Following previous works \cite{VINVL,LEMON,BLIP,SimVLM}, we also adopt the large-scale pre-training on the noisy image-text corpus to improve the downstream captioning task.
We combine three widely used pre-training losses including the language modeling (LM) loss, image-text contrastive loss \cite{CLIP,ALIGN}, and image-text matching loss \cite{UNITER,ALBEF,BLIP} to jointly optimize the visual encoder and cross-modal fusion model, as follows:
\begin{equation}\label{caption loss}
	{\cal L}_{\text{Pre-train}} = {\cal L}_{\text{Contrast}} + {\cal L}_{\text{Match}} + {\cal L}_{\text{LM}}.
\end{equation}
Similar to the CLIP \cite{CLIP} and ALIGN \cite{ALIGN}, the contrastive loss ${\cal L}_{\text{Contrast}}$ measures the similarity of the image-text pairs via a late fusion manner such as dot-product. 
This loss aims to align the feature representations of the visual input and text input by encouraging positive
image-text pairs to have similar representations in contrast to the negative pairs.
The matching loss ${\cal L}_{\text{Match}}$ measures the image-text similarity via a deep cross-modal fusion manner such as cross-attention.
Then, the output embedding of the special token \texttt{[CLS]} is used to conduct the binary classification to judge whether an image-text pair is positive (matched) or negative (unmatched).
The language modeling loss ${\cal L}_{\text{LM}}$ optimizes a cross-entropy loss which trains the model to maximize the likelihood of the text in an autoregressive manner.
It has been widely recognized that the aforementioned losses can facilitate cross-modal alignment \cite{BLIP}.
Therefore, although we focus on the image captioning, we additionally include the ${\cal L}_{\text{Contrast}} $ and ${\cal L}_{\text{Match}} $ in the pre-training stage.
For more details, please refer to BLIP~\cite{BLIP}.

\begin{table}[h]
	\scriptsize
	\centering
	\begin{tabular*}{8.5 cm} {@{\extracolsep{\fill}}l|c}
		\hline
		~ Setting &Approximate Prompt Tokens \\
		\hline
		~ \multirow{2}{*}{COCO Prompt} & \texttt{toned extend blank turnissbyzone shells lady}  \\
		&\texttt{elderea tagwing painted phone electronics} \\
		\hline
		~ \multirow{2}{*}{TextCap Prompt} & \texttt{edit extend pierce turnissbyzone shellsali}  \\
		&\texttt{elderea tagzo painted phone fuel} \\
		\hline
	\end{tabular*}
	\caption{Examples of the approximate auto-prompts via searching for the words that are closest to the learned prompt embeddings according to the dot-product similarity. } \label{table: learned prompts}	
\end{table}

\begin{table}[h]
	\scriptsize
	\centering
	\begin{tabular*}{8.5 cm} {@{\extracolsep{\fill}}l|c}
		\hline
		~  & Emotional Words \\
		\hline
		~ \multirow{2}{*}{Pre-defined Positive Words} & \texttt{happy}, \texttt{nice}, \texttt{awesome}, \texttt{tasty}, \texttt{great}  \\
		&\texttt{pretty}, \texttt{beautiful}, \texttt{cute}, \texttt{good}, \texttt{delilcious}\\
		\hline
		~ \multirow{2}{*}{Pre-defined Negative Words} & \texttt{stupid}, \texttt{bad}, \texttt{lonely}, \texttt{disgusting}, \texttt{silly}  \\
		&\texttt{dead}, \texttt{ugly}, \texttt{crazy}, \texttt{terrible}, \texttt{dirty} \\
		\hline
	\end{tabular*}
	\caption{Details of the pre-defined emotional words. } \label{table: pos and neg words}	
\end{table}

\begin{table}[h]
	\scriptsize
	\centering
	\begin{tabular*}{8.0 cm} {@{\extracolsep{\fill}}l|cccccc}
		\hline
		~ &COCO  &VG  &SBU  &CC3M  &CC12M &LAION  \\
		\hline
		~ Image &113K &100K &860K &3M &10M &115M  \\
		~ Text  &567K &769K &860K &3M &10M &115M  \\
		\hline
	\end{tabular*}
	\caption{Details of the pre-training datasets. } \label{table: pretrain_data}	
\end{table}

\begin{table*}[h]
	\scriptsize
	\centering
	\begin{tabular*}{12.0 cm} {@{\extracolsep{\fill}}l|cccc|cccc}
		\hline
		~  \multirow{2}{*}{Method} &\multicolumn{4}{c|}{Validation Set}  &\multicolumn{4}{c}{Test Set} \\
		~   &B@4 &M &C &S &B@4 &M &C &S  \\
		\hline
		~   BUTD \cite{BUTD} &20.1  &17.8 &41.9 &11.7 &14.9 &15.2 &33.8 &8.8 \\
		~   AoANet \cite{AoANet} &20.4  &18.9 &42.7 &13.2 &15.9 &16.6 &34.6 &10.5 \\
		~   M4C \cite{M4C} &23.3 &22.0 &89.6 &15.6 &18.9 &19.8 &81.0 &12.8 \\
		~   M4C (GT OCR) \cite{M4C}  &26.0 &23.2 &104.3 &16.2 &21.3 &21.1 &97.2 &13.5 \\
		~   MMA-SR \cite{MMASR} &24.6 &23.0 &98.0 &16.2 &19.8 &20.6 &88.0 &13.2 \\ 
		~   CNMT \cite{CNMT}  &24.8 &23.0  &101.7 &16.3 &20.0 &20.8 &93.0 &13.4 \\
		~   TAP \cite{TAP}  &25.8 &23.8 &109.2 &17.1 &21.9 &21.8 &103.2 &14.6 \\ 
		~   {\bf ConCap (Ours)} &{\bf 31.3} &{\bf 26.0} &{\bf 116.7} &{\bf 19.6} &{\bf 27.4} &{\bf 23.9} &{\bf 105.6} &{\bf 17.3} \\ 
		\hline
	\end{tabular*}
	\caption{Comparison results on the TextCaps \cite{textcaps} validation set and test set.} \label{table:textcap_full_results}	
\end{table*}

\begin{table}[h]
	\scriptsize
	\centering
	\begin{tabular*}{8.5 cm} {@{\extracolsep{\fill}}l|cccc}
		\hline
		~ Manual Prompt  &B@4 &M &C &S  \\
		\hline
		~ \texttt{a picture of} &23.7 &21.2 &83.8 &15.8  \\
		~ \texttt{a photo of}  &25.3 &22.0 &88.3 &16.4  \\
		~ \texttt{a picture contains}  &27.3 &22.8 &92.5 &18.0  \\
		~ \texttt{a picture with}  &22.3 &20.2 &79.2 &15.0 \\
		~ \texttt{a picture that shows}   &24.5 &22.3 &88.2 &16.4  \\
		~ \texttt{a beautiful picture that shows} &11.7 &15.9 &58.5 &11.4\\
		~ \texttt{a terrible picture that shows} &17.8 &19.0 &71.9 &14.1  \\
		~ \texttt{a normal picture that shows} &19.7 &19.7 &76.1 &14.7  \\
		\hline
	\end{tabular*}
	\caption{Performance of different manual prompts on the COCO Karpathy test split \cite{MSCOCO}.} \label{table: manual prompts on COCO}	
\end{table}

{\bf \noindent Inference Stage.} In the inference stage, we use beam search with a beam size of 3 to generate captions.
For COCO-style, TextCap-style, Positive, Negative, and Short-length captions, the maximum generation length is set to 40 since the learned prompts have already occupied 16 tokens.
For Medium-length and High-length captions, the maximum generation length is set to 60.

{\bf \noindent Training Data.} To divide the training data by different emotions, we select the following 10 positive words and 10 negative words, as shown in Table~\ref{table: pos and neg words}.
Note that the total positive and negative captions in the COCO dataset \cite{MSCOCO} are rare (only 1.5\% of the total COCO dataset).
Prompt learning facilitates the \emph{few-shot} domain transfer using limited training samples.

Table~\ref{table: pretrain_data} shows the details of the pre-training datasets.

\section{Detailed Results on TextCaps}
Due to the limited space, in the main paper, we only exhibit the BLEU@4 and CIDEr results on the TextCaps dataset \cite{textcaps}.
The compact comparison results are shown in Table~\ref{table:textcap_full_results}.

\section{Experiments on Manual Prompt}

In this section, we exhibit the captioning performance of different manual prompts.
To avoid the time-consuming model training, we evaluate the pre-trained model on downstream dataset COCO \cite{MSCOCO} without fine-tuning (i.e., zero-shot setting).
As shown in Table~\ref{table: manual prompts on COCO}, we validate the zero-shot performance with different manual prompts such as ``\texttt{a picture of}'', ``\texttt{a picture contains}'', ``\texttt{a photo of}'', ``\texttt{a picture that shows}'', etc.
From the results, we can observe that a slight word change will cause a clear positive or negative impact on the captioning performance.
To ensure each dataset has a specific manual prompt, we cannot utilize the common prompts such as ``\texttt{a picture of}''.
To this end, we heuristically design ``\texttt{a normal picture that shows}'' and ``\texttt{a textual picture that shows}'' for COCO and TextCaps datasets, respectively.

\section{More Visualization Results}

\begin{figure*}[h]
	\centering
	\includegraphics[width=17.3cm]{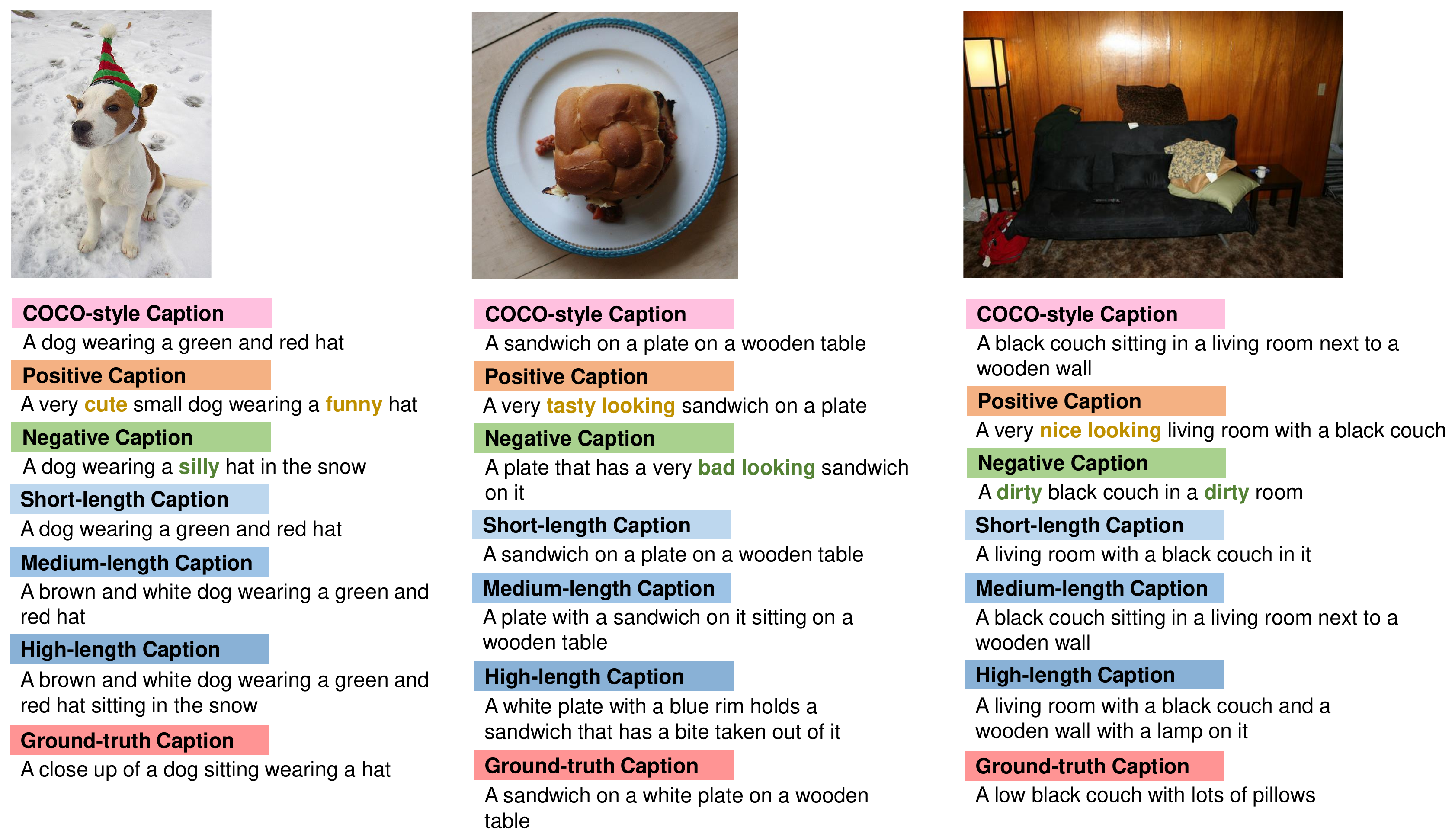}
	\includegraphics[width=17.3cm]{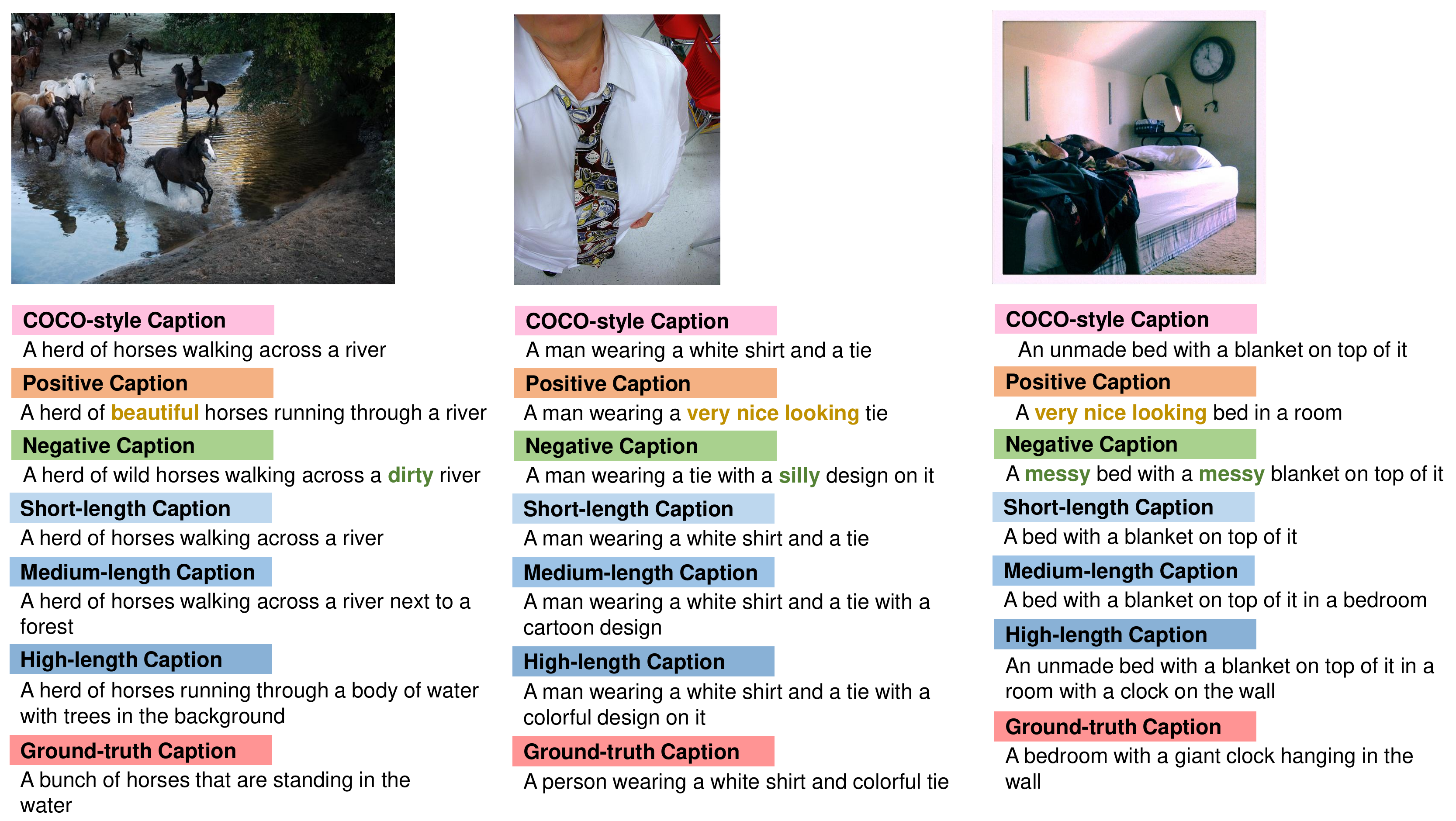}
	\caption{Image captioning examples from COCO \cite{karpathy} with different styles including COCO-style [$ \color[RGB]{255,193,224} \blacksquare$], Positive [$ \color[RGB]{244,177,131} \blacksquare$], Negative [$ \color[RGB]{169,209,142} \blacksquare$], Short-length [$ \color[RGB]{189,215,238} \blacksquare$], Medium-length [$ \color[RGB]{157,195,230} \blacksquare$], and High-length [$ \color[RGB]{138,178,214} \blacksquare$].}
	\label{fig:supp_coco_example}
\end{figure*}

\begin{figure*}[h]
	\centering
	\includegraphics[width=17.3cm]{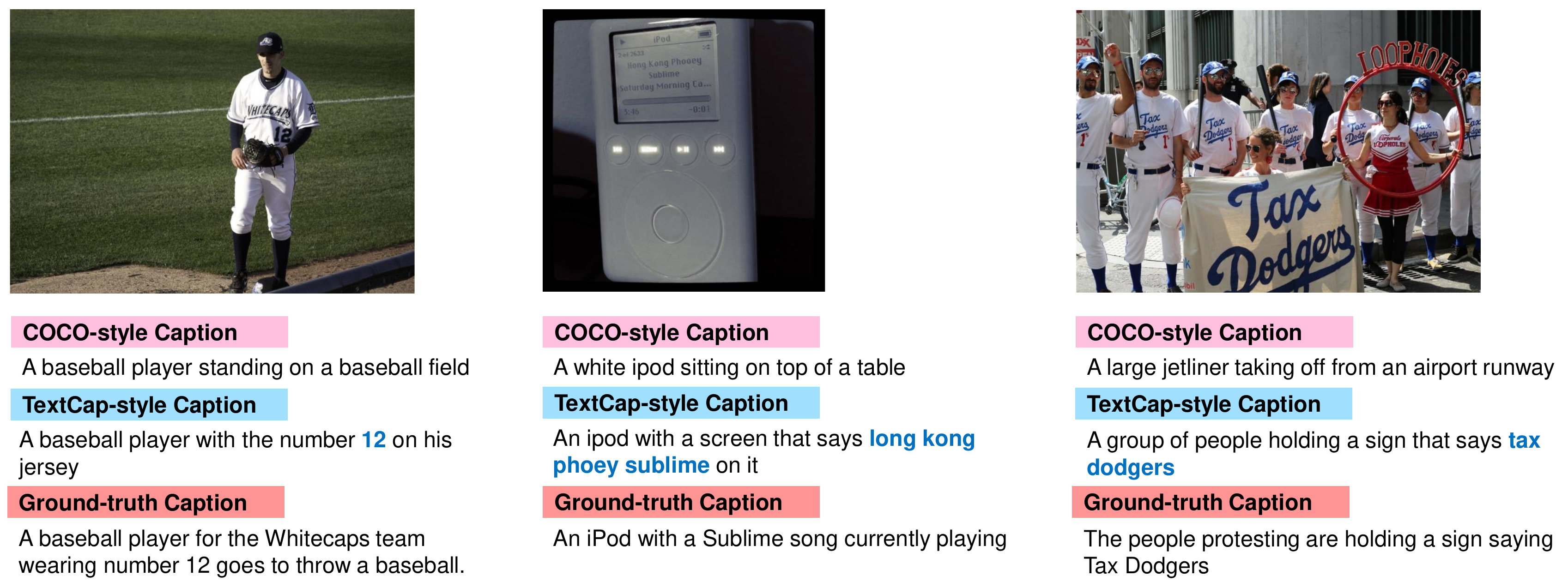}
	\includegraphics[width=17.3cm]{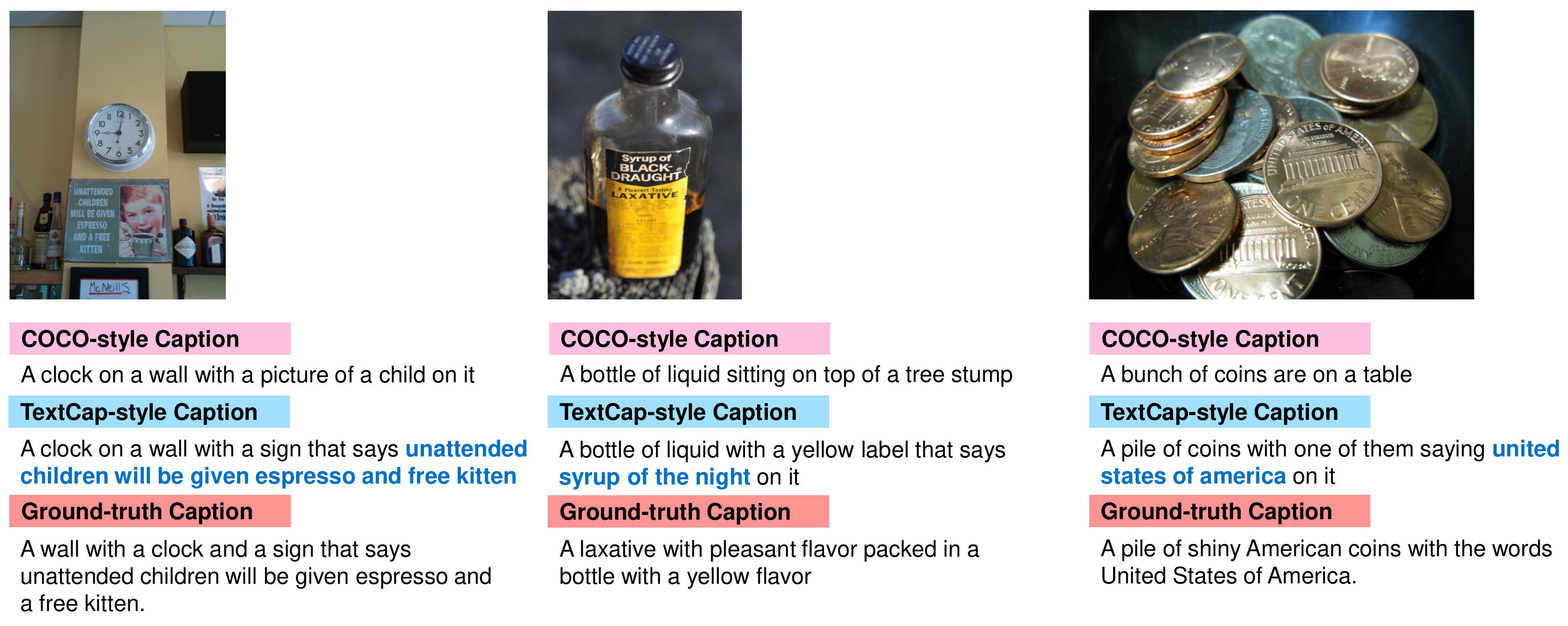}
	\includegraphics[width=17.3cm]{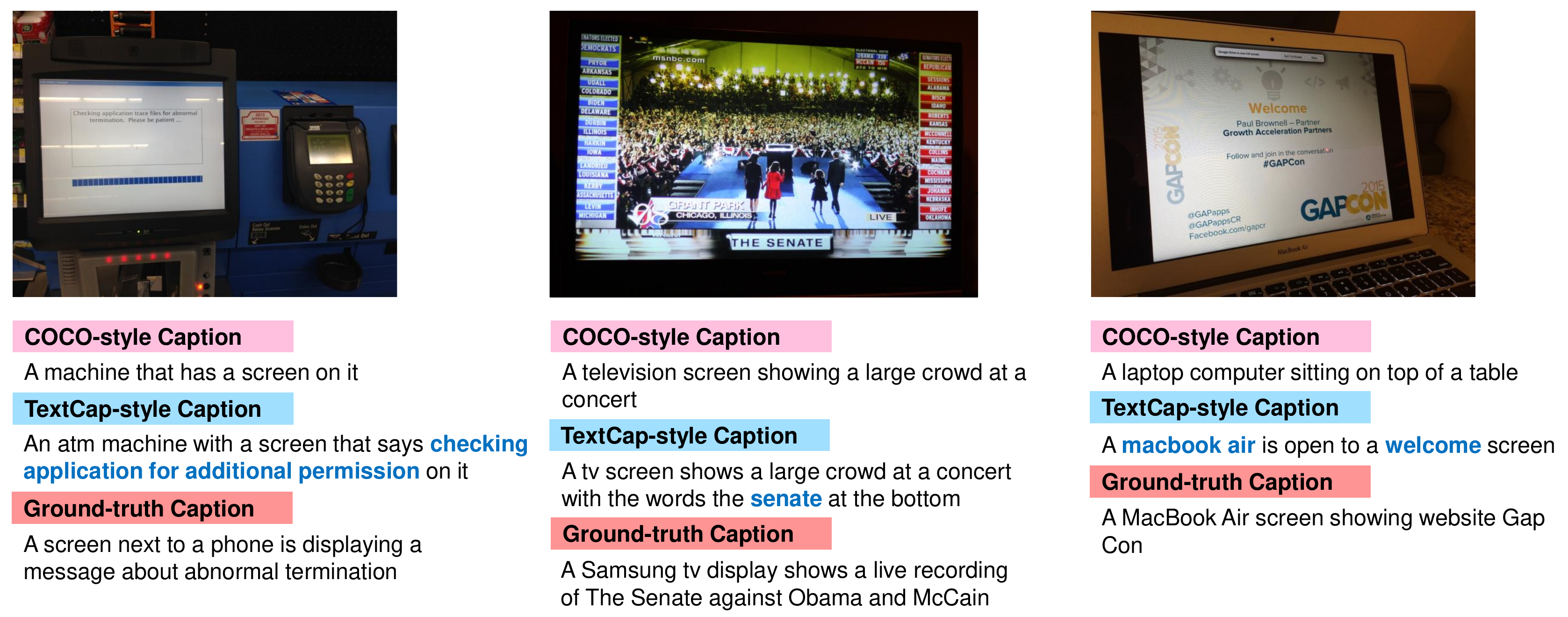}
	\caption{Image captioning examples from TextCaps validation split \cite{textcaps} with different styles including COCO-style [$ \color[RGB]{255,193,224} \blacksquare$] and TexCap-style [$ \color[RGB]{161,224,255} \blacksquare$]. Best view in color and zoom in.}
	\label{fig:supp_textcap_example}
\end{figure*}

In Figure~\ref{fig:supp_coco_example} and Figure~\ref{fig:supp_textcap_example}, we exhibit more visualization results on the COCO \cite{MSCOCO} and TextCaps \cite{textcaps} datasets, respectively.

To split the training data by different emotions, we pre-define some positive and negative words, as illustrated in Table~\ref{table: pos and neg words}.
Interestingly, our approach can generalize to more emotional words such as ``\texttt{funny}'' and ``\texttt{messy}'' that are not contained in Table~\ref{table: pos and neg words}. 
The emotional captioning results are shown in Figure~\ref{fig:supp_coco_example}.
These visualization results justify that prompt learning facilitates the few-shot learning with limited training samples.

\end{document}